\documentclass[runningheads]{llncs}
\usepackage[T1]{fontenc}
\usepackage{graphicx,verbatim}
\usepackage{amsmath,amssymb}
\usepackage{svg}

\usepackage{booktabs}
\usepackage{multirow}
\usepackage{makecell}
\usepackage{graphicx}
\begin{document}
\title{MAMA-FLUX.2: Image-to-Image Synthesis of Post-Contrast Breast DCE-MRI for the MAMA-SYNTH Challenge}
\titlerunning{MAMA-FLUX.2}
%
\author{Kamil Kwarciak\inst{1}\orcidID{0000-0002-1392-4291} \and
Marek Wodzinski\inst{1,2}\orcidID{0000-0002-8076-6246}}
\authorrunning{K. Kwarciak and M. Wodzinski}
%
\institute{Department of Measurement and Electronics, AGH University of Krakow, Krakow, Poland \and
Sano Centre for Computational Medicine, Krakow, Poland
\email{\{kwarciak,wodzinski\}@agh.edu.pl}}


  
\maketitle              
\begin{abstract}
Dynamic contrast-enhanced breast MRI is central to cancer diagnosis and monitoring, but requires gadolinium-based contrast agents. In this work, we address pre-to-post contrast breast MRI synthesis for the MAMA-SYNTH challenge. We propose MAMA-FLUX.2, a conditional latent flow-matching approach based on FLUX.2-Klein-4B. The pre-contrast image is encoded as spatial conditioning, while the model predicts the flow field associated with the post-contrast target latent. To adapt the pretrained model efficiently, we use LoRA fine-tuning and introduce a regional training objective combining global flow matching, tumor-region supervision, and stable foreground regularization. We further investigate LoRA rank, intensity windowing, and regional loss weights on axial slices, prioritizing clinically relevant tumor-focused metrics. Our ablation study shows that moderate tumor and stable-foreground weighting improves the
trade-off between image fidelity and tumor-region accuracy. The final model achieves the best overall balance with LoRA rank/$\alpha=64/64$,
$\mathrm{MHA}_{\max}=25$, $\lambda_{\mathrm{tumor}}=0.25$, and
$\lambda_{\mathrm{stable}}=0.1$. These results demonstrate that compact pretrained rectified-flow transformers can be adapted for contrast-enhanced MRI
synthesis using parameter-efficient fine-tuning and task-aware regional losses.


\end{abstract}

\section{Introduction}

Dynamic contrast-enhanced magnetic resonance imaging (DCE-MRI) is central to breast cancer imaging, as contrast uptake provides clinically important information about lesion conspicuity, morphology, and enhancement behaviour. However, post-contrast imaging requires intravenous gadolinium administration, increasing examination complexity and patient burden and limiting use in some clinical settings. This has motivated interest in virtual contrast enhancement, where post-contrast images are synthesized from pre-contrast acquisitions.

The MAMA-SYNTH Challenge formalizes this task by evaluating methods that synthesize peak-enhanced post-contrast breast MR images from corresponding pre-contrast inputs~\cite{osuala2026mamasynth}. The task is challenging because synthesized images must show plausible enhancement while preserving patient-specific anatomy. Since clinically relevant enhancement is often localized to small tumor regions, global image-level objectives alone may be insufficient. Effective models must therefore produce realistic images, preserve tumor-specific enhancement patterns, and maintain anatomical structures relevant to segmentation and image-level evaluation.

Recent diffusion and flow-based generative models have shown strong performance in high-fidelity image synthesis and image-to-image translation. Latent generative modelling improves efficiency by operating in a compressed representation rather than pixel space~\cite{rombach2022high}, while flow matching and rectified-flow formulations provide an effective framework for learning transformations between noise and data distributions~\cite{lipman2022flow,liu2022flow}. These properties make latent rectified-flow models well suited to conditional medical image synthesis, where realism and structural consistency are both required.

In this work, we adapt FLUX.2-Klein-4B, a compact latent rectified-flow transformer backbone~\cite{blackforestlabs2026flux2klein4b}, to pre-to-post contrast breast MRI synthesis. We formulate the task as conditional latent rectified-flow modeling: the pre-contrast image is encoded as spatial conditioning, while the model predicts the flow field for a noised post-contrast target latent. To better match the clinical structure of the task, we introduce a three-component objective combining global latent flow matching, tumor-region supervision, and stable-foreground preservation. This encourages accurate contrast synthesis while discouraging unnecessary changes in non-tumor foreground anatomy.

\textbf{Contributions.}
Our contributions are threefold. First, we adapt a pretrained latent rectified-flow transformer to conditional pre-to-post contrast breast MRI synthesis. Second, we introduce a tumor-aware and anatomy-preserving training objective that combines global latent flow matching with regional supervision. Third, we evaluate the effects of LoRA rank, LoRA scaling, and regional loss weighting through an ablation study using both image-quality and tumor-focused metrics.

\section{Methods}

We formulate pre-to-post contrast synthesis as conditional atent rectified-flow modeling~\cite{lipman2022flow,liu2022flow}. Our method builds on FLUX.2-Klein-4B~\cite{blackforestlabs2026flux2klein4b}, a compact rectified-flow transformer backbone inspired by recent high-resolution rectified-flow transformers~\cite{esser2024scaling}. Following the latent diffusion paradigm~\cite{rombach2022high}, we operate in the VAE latent space and adapt the model to the MAMA-SYNTH task of synthesizing peak-enhanced post-contrast breast DCE-MRI from pre-contrast input~\cite{osuala2026mamasynth}.

\subsection{Conditional Latent Flow Matching}

Let $x_0$ denote the pre-contrast conditioning image and $y_1$ the corresponding clean post-contrast target image. Their VAE latents are given by
$z_0 = E(x_0)$, and $z_1 = E(y_1)$,
where $E(\cdot)$ denotes the VAE encoder. Given Gaussian noise $\epsilon \sim \mathcal{N}(0,I)$ and a sampled noise level $t \in [0,1]$, we construct the noised target latent as
\begin{equation}
z_t = (1-t)z_1 + t\epsilon .
\end{equation}
Thus, $t=0$ corresponds to the clean target latent, while $t=1$ corresponds to pure noise. The corresponding flow-matching target is
\begin{equation}
u = \epsilon - z_1 .
\end{equation}
Equivalently, since $\epsilon = z_1 + u$, we have
\begin{equation}
z_t = z_1 + tu,
\end{equation}
and therefore
\begin{equation}
z_1 = z_t - tu .
\end{equation}

The rectified-flow transformer is conditioned on the pre-contrast latent $z_0$, the text embedding $c$, and the timestep embedding, and predicts the flow field
\begin{equation}
\hat{u}_{\theta} = f_{\theta}(z_t, z_0, c, t).
\end{equation}
The corresponding estimate of the clean post-contrast latent is
\begin{equation}
\hat{z}_1 = z_t - t\hat{u}_{\theta},
\end{equation}
and the synthetic post-contrast image is reconstructed by VAE decoding,
\begin{equation}
\hat{y}_1 = D(\hat{z}_1),
\end{equation}
where $D(\cdot)$ denotes the VAE decoder.

\subsection{Tumor-Aware and Anatomy-Preserving Objective}

We train the model using a three-component objective. The first component is the standard global flow-matching loss. For latent element $k$, corresponding to a channel and spatial position in the latent-error tensor, the squared flow-matching error is
\begin{equation}
e_k =
\left(
\hat{u}_{\theta,k} - u_k
\right)^2 .
\end{equation}
The global loss is defined as
\begin{equation}
\mathcal{L}_{\mathrm{global}}
=
\frac{1}{N}
\sum_{k=1}^{N}
e_k ,
\end{equation}
where $N$ is the total number of latent elements.

The second component uses the available tumor mask to emphasize accurate synthesis within the lesion region. Let $m^T_k$ denote the tumor mask resized to the latent-error resolution and broadcast over latent channels. The tumor-region loss is
\begin{equation}
\mathcal{L}_{\mathrm{tumor}}
=
\frac{
\sum_{k=1}^{N} m^T_k e_k
}{
\sum_{k=1}^{N} m^T_k
} .
\end{equation}

The third component encourages preservation of stable foreground anatomy outside the tumor. We define an image-space stable-foreground mask as
\begin{equation}
m^S =
\mathbf{1}[m^T = 0]\,
\mathbf{1}[x_0 \geq \tau_{\mathrm{fg}}]\,
\mathbf{1}[y_1 - x_0 \leq \tau_{\mathrm{enh}}],
\end{equation}
where $\tau_{\mathrm{fg}}$ suppresses background regions and $\tau_{\mathrm{enh}}$ excludes strongly enhancing tissue. We set these thresholds empirically to $\tau_{\mathrm{fg}} = -0.45$ and $\tau_{\mathrm{enh}} = 0.5$. After resizing this mask to the latent-error resolution and broadcasting it over latent channels, we obtain $m^S_k$. The stable-foreground loss is
\begin{equation}
\mathcal{L}_{\mathrm{stable}}
=
\frac{
\sum_{k=1}^{N} m^S_k e_k
}{
\sum_{k=1}^{N} m^S_k
} .
\end{equation}

The final training objective is
\begin{equation}
\mathcal{L}
=
\mathcal{L}_{\mathrm{global}}
+
\lambda_{\mathrm{tumor}} \mathcal{L}_{\mathrm{tumor}}
+
\lambda_{\mathrm{stable}} \mathcal{L}_{\mathrm{stable}} ,
\end{equation}
where $\lambda_{\mathrm{tumor}}$ and $\lambda_{\mathrm{stable}}$ control the relative contributions of the tumor-region and stable-foreground objectives, respectively.

\section{Experiments}

\subsection{Dataset}

We use MAMA-SYNTH challenge data, derived from the MAMA-MIA breast DCE-MRI dataset~\cite{garrucho2025large}. Each sample contains a paired pre-contrast image and peak-enhanced post-contrast target. Images are provided as 2D z-score-normalized MHA slices, with tumor masks for regional supervision and evaluation.

\subsection{Preprocessing}

We follow the official MAMA-SYNTH preprocessing protocol to convert DCE-MRI volumes into paired 2D slices. For each case, the pre-contrast acquisition is used as the conditioning input, while the peak-enhanced post-contrast slice is used as the synthesis target. All slices are resampled and stored as 2D MHA images, with spatial metadata preserved for evaluation. Image intensities are z-score normalized using the dataset-level pre-contrast statistics provided by the challenge. During training and inference, normalized MHA intensities are linearly clipped and mapped to the image range expected by the pretrained latent rectified-flow model. Unless otherwise stated, we use $[-0.51, 25]$ as the z-score window for the final model, while alternative upper bounds are evaluated in the ablation study. Since the pretrained model expects three-channel inputs, each scalar MHA slice is replicated across the RGB channels before VAE encoding. After decoding, the prediction is converted back to a single-channel normalized MHA slice.

\subsection{Experimental Setup}

\begin{figure}[b]
\centering
\includegraphics[width=\linewidth]{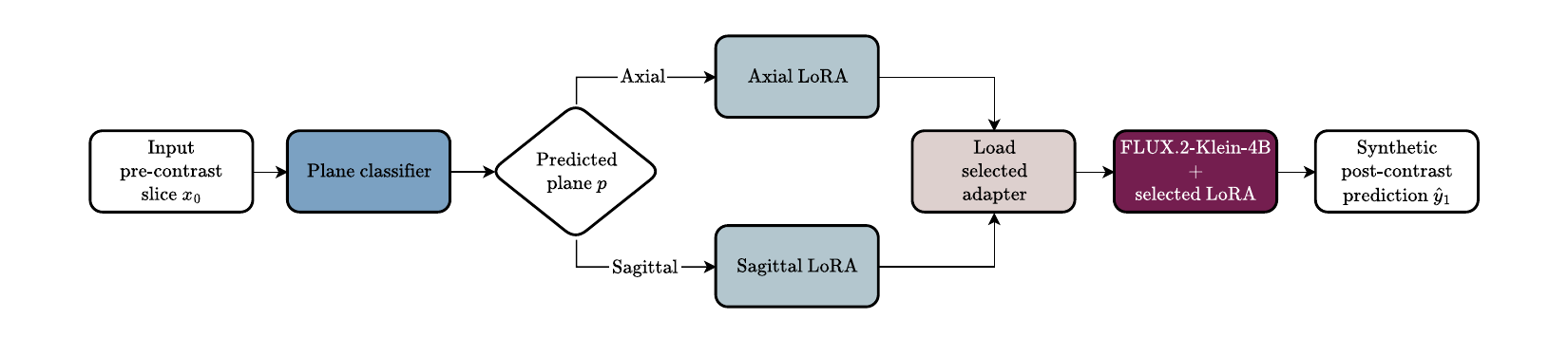}
\caption{
Plane-aware inference routing. A lightweight binary classifier predicts the anatomical plane of the input pre-contrast slice and selects the corresponding axial or sagittal LoRA adapter. The selected adapter is then used by the frozen FLUX.2-Klein-4B backbone for conditional post-contrast synthesis.
}
\label{fig:routing}
\end{figure}

\subsubsection{Anatomical Plane Routing}

Initial experiments suggested that plane-specific adaptation improves synthesis quality. We therefore train separate LoRA adapters for axial and sagittal slices. Although the MAMA-SYNTH validation and test sets are expected to contain axial slices, this routing mechanism supports both acquisition planes (Figure \ref{fig:routing}). A lightweight binary logistic-regression classifier predicts the anatomical plane of each pre-contrast slice and selects the corresponding adapter for inference. Image synthesis is then performed by the FLUX.2-Klein-4B backbone conditioned on the input image. The classifier uses hand-crafted features extracted from the 2D MHA slices, including image dimensions, intensity statistics and percentiles, threshold-based pixel fractions, foreground shape descriptors, and coarse row- and column-wise occupancy profiles. The foreground mask is defined as $x_0 \geq \tau_{\mathrm{fg}}$, with $\tau_{\mathrm{fg}}=-0.45$, consistent with the stable-foreground loss. Features are standardized, and class imbalance is addressed using balanced logistic-regression weights.

\subsubsection{FLUX.2 Fine-Tuning}

\begin{figure}[t]
\centering
\includegraphics[width=\linewidth]{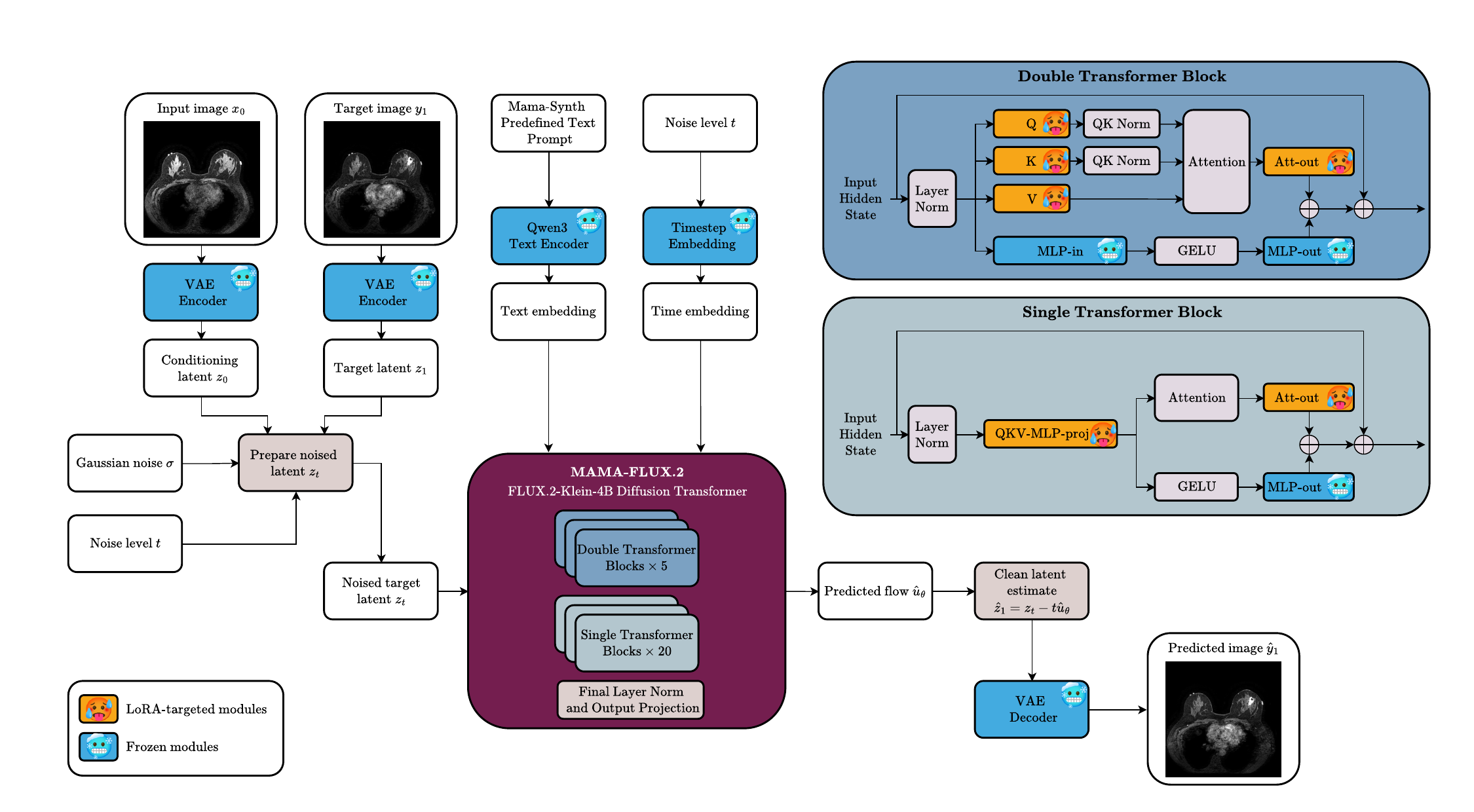}
\caption{
Overview of the proposed MAMA-FLUX.2 fine-tuning pipeline. The pre-contrast slice is encoded into the VAE latent space and used as spatial conditioning for a frozen FLUX.2-Klein-4B rectified-flow transformer. Only LoRA adapters inserted into the attention projections are optimized, while the VAE, text encoder, and pretrained transformer weights remain frozen.
}
\label{fig:mama-flux2-lora}
\end{figure}

We fine-tune FLUX.2-Klein-4B using low-rank adaptation (LoRA), while keeping the VAE, text encoder, and pretrained transformer backbone frozen. LoRA layers are inserted into the main attention projections of the transformer, including the query, key, value, and output projections of the double transformer blocks, as well as the joint projection and attention output projections of the single transformer blocks. This restricts optimization to a small set of adapter parameters while preserving the pretrained generative prior. The LoRA fine-tuning procedure is presented in Figure \ref{fig:mama-flux2-lora}. For all fine-tuning runs, we use the fixed text prompt: \emph{``Synthesize the matching peak-enhancement post-contrast breast DCE-MRI slice from this pre-contrast breast DCE-MRI slice. Preserve anatomy, acquisition plane, lesion location, and keep non-breast background dark.``} We perform the main hyperparameter search on axial slices, since the challenge validation and test sets are indicated to be axial. Based on these experiments, we transfer the selected configuration to the sagittal setting. The ablation study varies the LoRA rank and scaling factor, using tied values $r/\alpha \in {16/16, 32/32, 64/64}$, the upper MHA intensity window $\mathrm{MHA}_{\max} \in {25,35,40}$, and the regional loss weights $\lambda_{\mathrm{tumor}}$ and $\lambda_{\mathrm{stable}}$. The final selected configuration uses $r=\alpha=64$, $\mathrm{MHA}{\max}=25$, $\lambda{\mathrm{tumor}}=0.25$, and $\lambda_{\mathrm{stable}}=0.1$, which provides the best overall trade-off between tumor-focused metrics and global image quality. This final adapter contains 66,846,720 trainable parameters, corresponding to 1.72\%  of the 3,875,544,576 parameters in the frozen base transformer. The resulting LoRA adapter occupies 255 MiB in memory.

Models are trained with AdamW~\cite{Loshchilov2017DecoupledWD}, a constant learning rate, bf16 mixed precision, FSDP, gradient checkpointing, a per-device batch size of 1, and gradient accumulation over 4 steps. This results in an effective batch size of 16 on four NVIDIA GH200 GPUs with 96 GB of VRAM each. At inference, we use 50 rectified-flow sampling steps with guidance scale 4.0. On the evaluation slices used for runtime profiling, with spatial sizes ranging from $256\times256$ to $512\times512$, the model call required 11.21 s per slice on average.

\section{Results}

\begin{table*}[t]
\centering
\caption{Ablation study of LoRA rank, MHA maximum value, and regional loss weights. Each row denotes one complete configuration. Results are reported as mean $\pm$ standard deviation. Best mean values are highlighted in bold.}
\label{tab:ablation_study}
\resizebox{\textwidth}{!}{%
\begin{tabular}{ccccccccc}
\toprule
\makecell{LoRA\\rank/$\alpha$} &
\makecell{MHA\\max} &
$\lambda_{\mathrm{tumor}}$ &
$\lambda_{\mathrm{stable}}$ &
DSC $\uparrow$ &
HD95 $\downarrow$ &
LPIPS $\downarrow$ &
MSE $\downarrow$ &
\makecell{Tumor\\SSIM $\uparrow$} \\
\midrule

16/16 & 35 & 0.25 & 0
& $0.8192 \pm 0.1212$ & $31.0522 \pm 62.1443$ & $0.0913 \pm 0.0256$ & $0.2956 \pm 0.4154$ & $0.7717 \pm 0.1192$ \\
16/16 & 35 & 1 & 0
& $0.7472 \pm 0.2565$ & $72.9320 \pm 147.1482$ & $0.1127 \pm 0.0274$ & $0.3858 \pm 0.6377$ & $0.8010 \pm 0.1045$ \\
16/16 & 35 & 2 & 0
& $0.7679 \pm 0.2075$ & $81.3741 \pm 148.7569$ & $0.1100 \pm 0.0307$ & $0.3547 \pm 0.5304$ & $0.7616 \pm 0.1882$ \\

\midrule

32/32 & 35 & 0.25 & 0
& $0.8436 \pm 0.1036$ & $28.8759 \pm 60.5567$ & $0.0857 \pm 0.0253$ & $0.3041 \pm 0.4552$ & $0.8169 \pm 0.0894$ \\
32/32 & 35 & 1 & 0
& $0.8110 \pm 0.2001$ & $58.5668 \pm 143.8272$ & $0.1016 \pm 0.0278$ & $0.3395 \pm 0.4910$ & $0.7869 \pm 0.1917$ \\
32/32 & 35 & 2 & 0
& $0.7890 \pm 0.2027$ & $61.1797 \pm 90.3439$ & $0.1118 \pm 0.0291$ & $0.3571 \pm 0.5071$ & $0.7999 \pm 0.1494$ \\

\addlinespace[0.25em]

32/32 & 25 & 0.25 & 0
& $0.8406 \pm 0.1221$ & $31.4182 \pm 61.5544$ & $0.0812 \pm 0.0269$ & $0.2897 \pm 0.4651$ & $0.8363 \pm 0.0891$ \\
32/32 & 25 & 1 & 0
& $0.8380 \pm 0.0980$ & $27.5710 \pm 53.3358$ & $0.1021 \pm 0.0288$ & $0.3384 \pm 0.4709$ & $0.8368 \pm 0.0776$ \\
32/32 & 25 & 2 & 0
& $0.8084 \pm 0.1190$ & $68.1819 \pm 90.7406$ & $0.1055 \pm 0.0307$ & $0.3437 \pm 0.4888$ & $0.8432 \pm 0.0708$ \\

\midrule

64/64 & 25 & 0.25 & 0.1
& $\mathbf{0.8722} \pm 0.0716$ & $\mathbf{16.8431} \pm 38.9849$ & $0.0745 \pm 0.0249$ & $0.2592 \pm 0.3837$ & $0.8610 \pm 0.0702$ \\
64/64 & 25 & 0.25 & 1
& $0.8263 \pm 0.1614$ & $33.2027 \pm 61.3941$ & $\mathbf{0.0710} \pm 0.0213$ & $\mathbf{0.2144} \pm 0.3506$ & $0.8672 \pm 0.0656$ \\
64/64 & 25 & 2 & 0
& $0.8632 \pm 0.0628$ & $28.3243 \pm 60.4912$ & $0.0764 \pm 0.0261$ & $0.2610 \pm 0.4015$ & $\mathbf{0.8748} \pm 0.0453$ \\

\addlinespace[0.25em]

64/64 & 35 & 0.25 & 0.25
& $0.8467 \pm 0.1023$ & $17.2244 \pm 39.2133$ & $0.0772 \pm 0.0208$ & $0.2460 \pm 0.3544$ & $0.8189 \pm 0.1472$ \\
64/64 & 35 & 0.25 & 0.1
& $0.8104 \pm 0.1986$ & $19.3245 \pm 39.7797$ & $0.0797 \pm 0.0238$ & $0.2667 \pm 0.3749$ & $0.8292 \pm 0.0932$ \\
64/64 & 35 & 0.1 & 0.25
& $0.8128 \pm 0.2031$ & $59.9237 \pm 145.0143$ & $0.0733 \pm 0.0219$ & $0.2276 \pm 0.3436$ & $0.7808 \pm 0.1656$ \\
64/64 & 35 & 0.1 & 0.1
& $0.7949 \pm 0.1722$ & $19.9224 \pm 39.2512$ & $0.0761 \pm 0.0198$ & $0.2267 \pm 0.3520$ & $0.8434 \pm 0.0618$ \\

\addlinespace[0.25em]

64/64 & 40 & 0.25 & 0.25
& $0.7621 \pm 0.2831$ & $54.3626 \pm 139.6625$ & $0.0853 \pm 0.0215$ & $0.2611 \pm 0.3673$ & $0.8201 \pm 0.1361$ \\

\bottomrule
\end{tabular}%
}
\end{table*}

\begin{figure}[t]
\centering
\includegraphics[width=\linewidth]{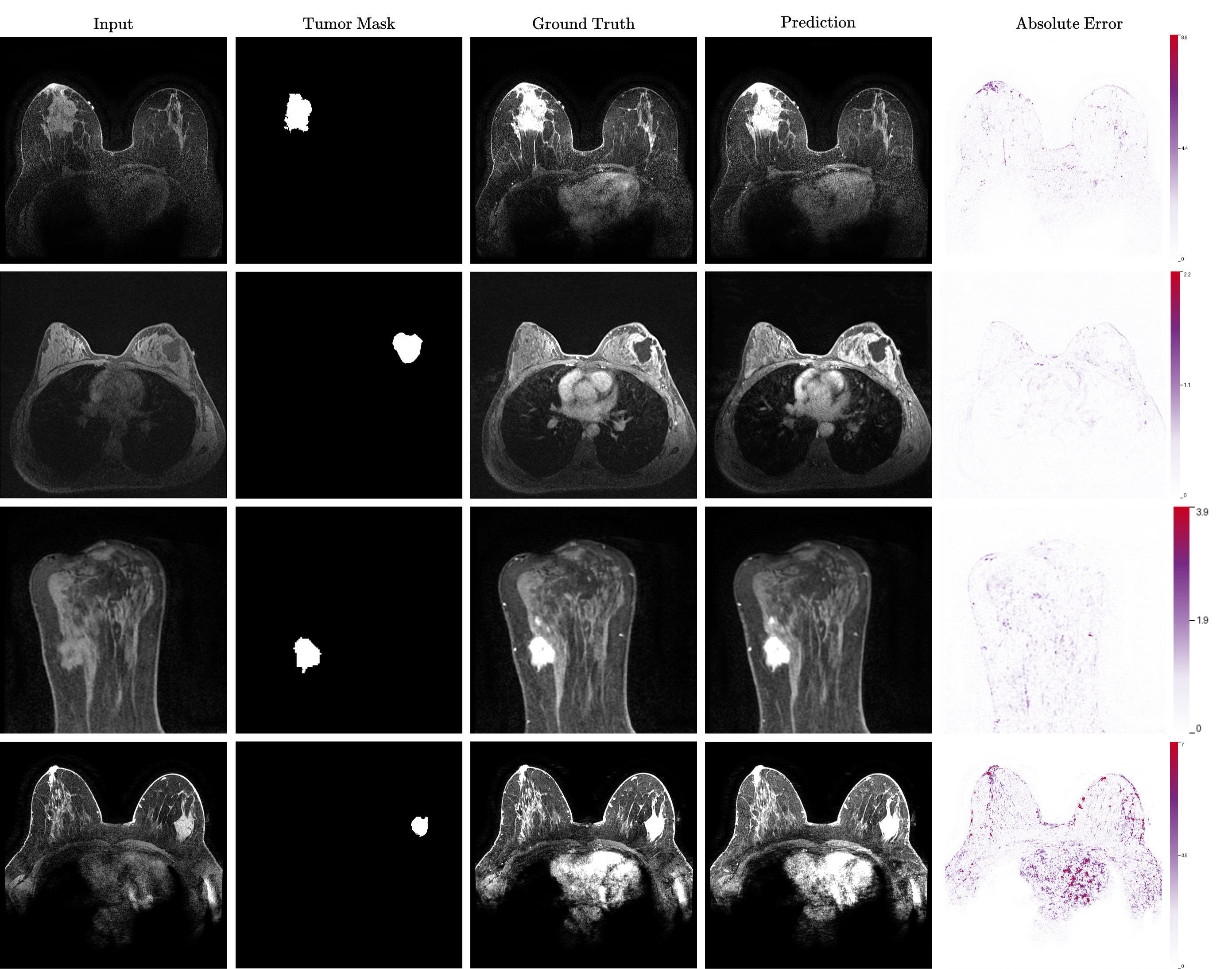}
\caption{Qualitative results of the selected model (LoRA rank/$\alpha=64/64$,
$\mathrm{MHA}_{\max}=25$, $\lambda_{\mathrm{tumor}}=0.25$,
$\lambda_{\mathrm{stable}}=0.1$). The predictions preserve the main anatomical structure and produce spatially aligned enhancement patterns in the shown examples. Residual errors are visualized with independently scaled error maps for each case.}
\label{fig:quality}
\end{figure}

We perform an exploratory hyperparameter analysis on axial slices, as the MAMA-SYNTH validation and test sets are reported to contain axial acquisitions. This analysis is conducted on a small development subset of 10 instances sampled from the available training data. Performance is evaluated using the Sørensen--Dice coefficient (DSC), 95th-percentile Hausdorff distance (HD95), tumor structural similarity index (tumor SSIM), mean squared error (MSE), and learned perceptual image patch similarity (LPIPS). The ablation study in Table~\ref{tab:ablation_study} suggests that moderate regional supervision provides the most favorable trade-off between global image fidelity and tumor-focused performance in this development setting. In particular, $\lambda_{\mathrm{tumor}}=0.25$ and $\lambda_{\mathrm{stable}}=0.1$ yield the strongest segmentation-oriented results, achieving the highest DSC and lowest HD95. Increasing the stable-foreground weight while keeping $\lambda_{\mathrm{tumor}}=0.25$ improves image-level metrics such as MSE and LPIPS, but degrades segmentation performance. Conversely, using a stronger tumor-region weight improves tumor ROI similarity in terms of tumor SSIM, but does not provide the best overall balance across metrics. Since the task requires preservation of localized enhancement patterns relevant to downstream tumor assessment, we prioritize DSC and HD95 over small gains in global image similarity. We therefore select the configuration with LoRA rank/$\alpha=64/64$, $\mathrm{MHA}*{\max}=25$, $\lambda*{\mathrm{tumor}}=0.25$, and $\lambda_{\mathrm{stable}}=0.1$ as the final setup. Qualitative results presented in Fig.~\ref{fig:quality} are broadly consistent with the quantitative trends. The selected model preserves the overall breast anatomy and produces spatially aligned enhancement patterns in most shown examples, while residual errors are most apparent in regions with strong or heterogeneous contrast uptake. To provide a fairer qualitative assessment, Fig.~\ref{fig:quality} also includes a case with visibly higher reconstruction error, illustrating that the model can still struggle with challenging enhancement patterns. Error maps indicate that discrepancies are concentrated primarily around enhancing tissue and tumor-adjacent regions rather than in the background. Overall, these exploratory results indicate that conditional latent rectified-flow modeling is a promising formulation for pre-to-post contrast breast MRI synthesis, particularly when combined with tumor-aware and anatomy-preserving supervision.

\section{Limitations}

This study is limited by the exploratory ablation on only 10 development cases, so the selected hyperparameters may not generalize to the hidden test set or external cohorts. The method is trained slice-wise and does not enforce volumetric consistency, while the intensity window, regional loss weights, and plane-routing classifier were selected empirically. These design choices should therefore be validated on larger, independent datasets and across different acquisition protocols.

\begin{credits}
\subsubsection{\ackname}We gratefully acknowledge Polish high-performance computing infrastructure PLGrid (HPC Center: ACK Cyfronet AGH) for providing computer facilities and support within computational grant no. PLG/2026/019392. This work was partially supported by the Excellence Initiative Research University program at the AGH University of Krakow.

\subsubsection{\discintname}
The authors have no competing interests to declare that are relevant to the content of this article.
\end{credits}

\bibliographystyle{splncs04}
\bibliography{mybibliography}
\end{document}